# Fine-Tuning Pre-trained Language Models to Detect In-Game Trash Talks

## Daniel Fesalbon[1], Arvin De La Cruz[2], Marvin Mallari[3], Nelson Rodelas[4]

[1,2]Graduate School, Polytechnic University of the Philippines
[3]School of Engineering, Computer and Library Studies, Holy Cross College, Philippines
[4]University of the East, Caloocan, Metro Manila, Philippines

**Abstract**
Common problems in playing online mobile and computer games were related to toxic behavior and abusive communication among players. Based on different reports and studies, the study also discusses the impact of online hate speech and toxicity on players' in-game performance and overall well-being. This study investigates the capability of pre-trained language models to classify or detect trash talk or toxic in-game messages. The study employs and evaluates the performance of pre-trained BERT and GPT language models in detecting toxicity within in-game chats. Using publicly available APIs, in-game chat data from DOTA 2 game matches were collected, processed, reviewed, and labeled as non-toxic, mild (toxicity), and toxic. The study was able to collect around two thousand in-game chats to train and test BERT (Base-uncased), BERT (Large-uncased), and GPT-3 models. Based on the three models' state-of-the-art performance, this study concludes pre-trained language models' promising potential for addressing online hate speech and in-game insulting trash talk.

## 1. Introduction

A report from 2016 highlighted the widespread nature of online harassment and abuse, indicating that 47% of internet users have encountered such behavior [1]. Within the gaming context, a study conducted by the UNITY, a gaming development platform, revealed that 72% of players had witnessed toxic behavior directed towards others while engaging in multiplayer video games [2]. These toxic behaviors encompass a range of actions, including sexual harassment, hate speech, threats of violence, and doxing, all of which can significantly impact individuals' in-game performance and mental health. Research has indicated that unhealthy dialogues during gameplay can disturb a player's concentration, enjoyment, and overall flow experience, particularly when these behaviors originate from within the gaming community itself [3]. Additionally, specific gaming genres like multiplayer online battle arena games are linked to negative effects on academic performance and interpersonal relationships, particularly in educational settings [4]. Toxic chats and negative comments at different online communities and platforms is rampant and has raised concerns about its impact on user experiences and personal well-being. This study aims to address the issue of toxicity in online gaming by utilizing and investigating pre-trained machine learning-based language models for the detection and mitigation of toxic chat messages and in-game profanity.

Different reporting systems currently exists in popular online games to address and limit toxic behavior. However, there is room for improvements. There are studies proposing additional measures to encourage positive conduct among players. One example is the recommendation system, which endorses players that exhibits good communication and team coordination [5]. Additionally, player feedback indicates that enhancing the detection of toxic behavior is vital in mitigating the problem [6]. Drawing on the insights and recommendations from these reports and studies [3, 4, 5, 6], this study explores the capability of pre-trained machine learning-based language models in detecting in-game toxic chats and messages. By utilizing these pre-trained language models, the study aims to contribute to the development of more effective ways to address toxic behavior within online gaming communities. An example of study which apply machine learning techniques to detect toxic behavior in gaming was from Blackburn et al. (2014). They discuss the negative impact of toxic behavior in competitive games. They focus on the game "League of Legends" (LoL) and its crowdsourcing platform called the Tribunal. The Tribunal is a two-stage system that utilizes player reports and human experts to judge and punish toxic players. However, this system is time-consuming and resource-intensive. They propose a solution using supervised learning to predict crowdsourced decisions on toxic behavior. They used a large dataset of over 10 million user reports involving 1.46 million toxic players and their corresponding crowdsourced decisions. The results indicate that the proposed approach performs well in identifying cases where there is an overwhelming majority consensus. The classifier's effectiveness is also demonstrated across different regions. Their study concludes by discussing the potential practical benefits of this approach, including cost savings and improved victim protection [16].

This study was inspired from toxic chat and behavior problems in online communities and gaming platforms. Using pre-trained machine learning-based language models, this study investigates the use of Bidirectional Encoder Representations from Transformers (BERT) [7] from Google researchers and Generative Pre-trained Transformer (GPT) [8] from OpenAI to detect toxic player in-game chats and aims to answer the following:
- The effectiveness of pre-trained machine-learning based language models in detecting in-game toxic chats and trash talks to address hate speech problems in different gaming platforms and online communities.
- Which models from BERT (Base-uncased), BERT (Large-uncased), and GPT-3 would display the best performance for the task of in-game chat toxicity detection.

## 2. Methods
### Data Acquisition and Processing
The dataset used in this study was collected from OpenDota's public API. OpenDota is a platform that offers free access to DOTA 2 data where users and developers can explore in-game data and utilize their available APIs to create their own applications using the data. The proponents developed a python script to collect data from the APIs to extract and obtain the necessary details from recorded games. The script creates one file every execution which contains the collected chat data. Iteratively, this step generates multiple files. Data from these then generated files were consolidated into one file dataset. The consolidated chat data were filtered using a language detector library in Python with only identified English chats to be included in the final dataset. This final dataset is employed for the fine-tuning and evaluation of the language models to the toxic chat detection task.

**Figure 1: Illustration of data gathering procedures conducted in this study to collect toxic chat data from the game DOTA 2**

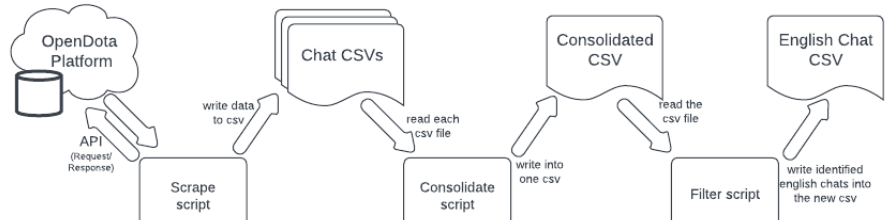

**Bidirectional Encoder Representations from Transformers (BERT)**

One language model that this study use was BERT [7]. It is a transformer-based [9] natural language processing (NLP) model developed by researchers from Google. Its pre-trained model can be fine-tuned on different NLP tasks such as text classification, sentiment analysis, and question answering. Transformer-based language models are notable with a key component called "attention mechanism" from the transformer architecture. This component, attention mechanism, assigns weights to each word based on its relevance to the other words in the order or sequence. It allows the model to understand the relationships between words in a statement or sentence. Unlike Recurrent Neural Networks (RNNs) [10], with attention, Transformer refers to multiple positions in a sequence for context at any given time regardless of distance. The BERT model consists of two versions, BERT-Base and BERT-Large. BERT-Large is a more computationally intensive model, but it has the ability to capture intricate patterns and dependencies in the input data, which contributes to achieving state-of-the-art results as reported in the original BERT paper. For this study, the proponents use the Base-uncased and Large-uncased versions of BERT.

**Generative Pre-trained Transformer (GPT)**

Another language model in this study will be GPT. GPT [8] is a language model developed by OpenAI. Similar to Google's BERT, GPT is also derived from the transformer architecture which utilizes self-attention mechanism enabling the model's ability to determine input element importance, making it ideal for language translation, text classification, and text generation [11]. It has been developed iteratively, with subsequent versions such as GPT-2 [12] and GPT-3 [13]. GPT has made significant development in the field of natural language processing (NLP), particularly with the release of GPT-3. These models are pre-trained on large amounts of internet text data. It is able to learn statistical patterns and semantic relationships between words and phrases making itself an excellent choice for NLP tasks such as summarization, text generation, sentence completion, and language translation. GPT's difference from BERT is it generates coherent text using autoregressive language modeling, whereas BERT uses masked language modeling to learn bidirectional representations. For this study, GPT version 3 (GPT-3) is used.

**Test Prediction Evaluation**

To evaluate the three models, the study's basis is Precision, Recall, F1-score, and Accuracy based on their predictions to the classification of the chats. Each of the metrics can be derived in the following formulas where TP = true positives, FP = false positives, FN = false negatives, and TN = true negatives.
- Precision measures the count of correct positive predictions.

- Recall measures the amount of the positive cases a model predicted correctly.
- F1-score is the measure combining both precision and recall.
- General Accuracy is the fraction of correct prediction or classification of a model.

## 3. Related Studies

The difference between GPT and BERT is that BERT processes text bidirectionally to capture context from both directions, while GPT is unidirectional and predicts the next word based on previous context. BERT is suited for natural language processing (NLP) tasks like language inference and question answering [7]. Whereas GPT has an advantage over NLP task like text and dialogue generation [18] and summarization. There are studies that conclude the superiority of BERT in different text classification tasks. One study was from Yadav et al. (2023) [17] where they examine how people perceive smart farming technologies by conducting sentiment analysis on data extracted from YouTube using 3 transformer models: GPT-2, BERT specifically fine-tuned for the agricultural domain, and a BERT variant, DistilBERT. The study's findings and statistical evaluations suggest that transformer models are effective for categorizing both technical and agricultural texts. Notably, the domain-specific pre-trained BERT model for agriculture demonstrates superior performance compared to the other two models which includes GPT-2. The evaluation metrics for the classification were the macro-F1-score (a metric that balances precision and recall) and accuracy. They also imply that transformers are a viable option for classifying technical and agricultural text. Another comparative study between BERT and GPT was from Rehana et al. (2023) [11] where they assessed the ability of various GPT and BERT models to identify protein-protein interactions (PPIs) using a benchmark dataset containing 164 PPIs within 77 sentences from the Learning Language in Logic (LLL) dataset. Though their findings suggests that GPT models are effective in identifying PPIs from text with the GPT-4 exhibited comparable performance to the best BERT models, they also conclude that in general, BERT-based models outperformed the GPT models. Few similarities with Rehana et al.'s approach, this study would add another conclusion in text classification between BERT and GPT by expanding the application of BERT and GPT to in-game communication. This study will be using two models from BERT, the BERT (Base-uncased) and BERT (Large-uncased), and one from GPT-3, the model called 'davinci-002'.

## 4. RESULTS

**Data Processing and Analysis**

Utilizing the OpenDota's API, the proponents gathered a corpus of approximately 11,074 chat messages through the implementation of the Python script. These chat messages underwent a classification process, wherein they were categorized based on their severity levels: non-toxic, mild, and toxic. The specific classification yielded 636 instances of 'toxic' chat messages and 524 instances of 'mild' toxic chat messages. The remaining messages were 'non-toxic'. In-depth analysis of the dataset revealed a notable imbalance among the severity classes. The study identified an imbalance in the dataset wherein according to Bach et al. (2019), an imbalanced dataset poses a challenge when the representation of each class does not constitute an equal part of dataset, leading to significant variations in the number of instances attributed to each class. In such cases, the classifier or predictive models might exhibit bias and produce inaccurate results [14]. In order to mitigate the implications of this imbalance, the proponents opted to employ an undersampling technique. This method involved the removal of 8,522 chat messages from the majority class (non-toxic), aligning its size more closely with that of the minor classes (mild and toxic chats). Consequently, the dataset used for the experiment was refined to comprise 2,552 instances, of which 1,392 instances were attributed to the 'non-toxic' category.

Table 1: Example of chats classified into non-toxic, mild, and toxic

| Non-toxic | Mild (toxic) | Toxic |
|---|---|---|
| "he scared" | "ez the fk" | "what a loser" |
| "not about it" | "are you some kind of retard bro" | "shut up, stupid boy" |
| "Took the kill" | "Wtf rs" | "fuck your mother aa" |
| "free to end" | "i ate a shit ton of broccoli" | "idiot this pudge" |
| "stop walking into my spells" | "you killed me and got 3 level wtf" | "why you tip me unskilled bitch(" |

Figure 2: Displays two visualized representation of chat classification counts from the original dataset (a) and the final dataset (b) which underwent undersampling

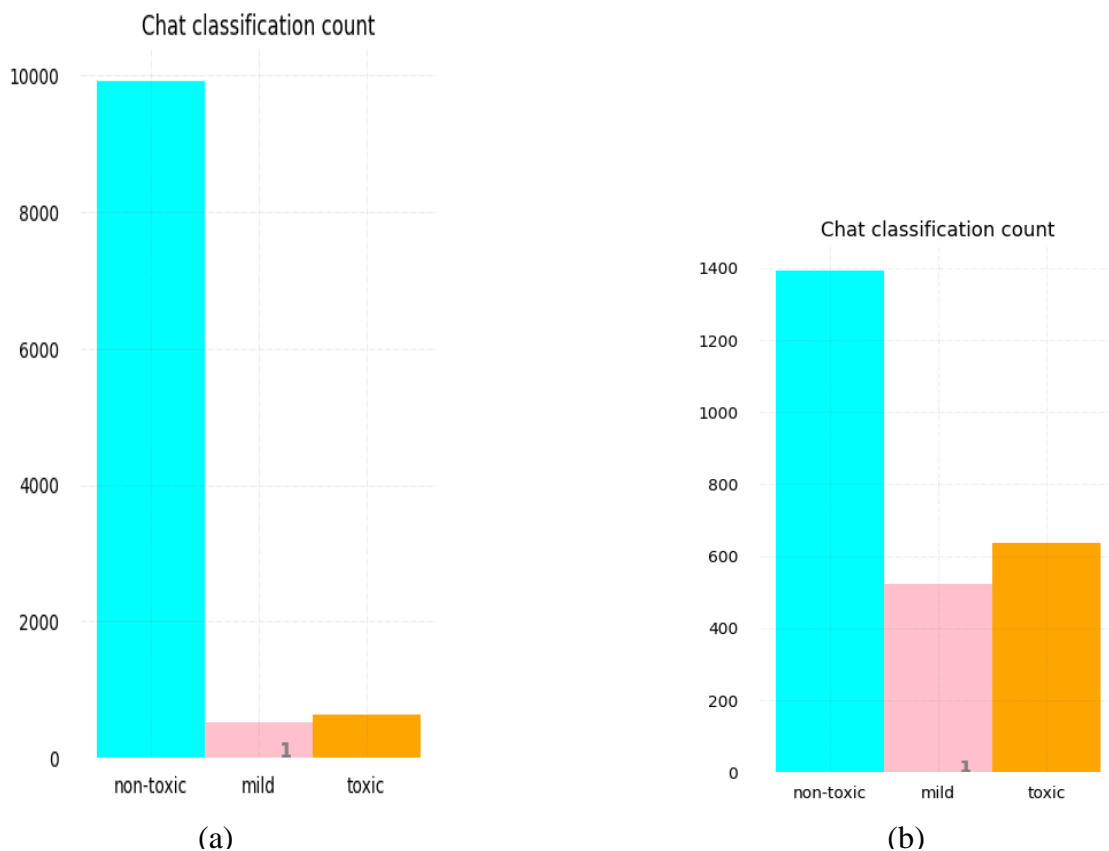

(a)                          (b)

This research study categorizes in-game chats into three distinct classifications according to the severity of toxicity. The first classification, referred to as 'non-toxic', pertains to chats that exhibit a standard conversational tone without any profane, insulting, offensive, or degrading language. The second classification, termed 'mild', refers chats containing profanity or inappropriate language, although these elements are not directed towards insulting or attacking any individual. The third classification, labeled as 'toxic', pertains to in-game chats that involve offensive, insulting, and discriminatory language aimed at engaging in derogatory trash-talking within the gaming context.

**Fine-tuning**

The investigation was conducted using HuggingFace's [15] python library for the BERT models and OpenAI's web API for the GPT-3 model. Pre-trained BERT (Base-uncased) and BERT (Large-uncased) were pulled from their open-source repository in the HuggingFace platform. The models were fine-tuned to the toxic chat message detection task.

**Table 2: Displays the list of parameters used by the BERT models in the fine-tuning task**

| Parameter | Value |
| --- | --- |
| Evaluation Strategy | Epoch |
| Number of Training Epochs | 5 |
| Training Batch Size | 16 |
| Evaluation Batch Size | 16 |
| Weight Decay | 0.01 |
| Scheduler | Linear |
| Optimizer | AdamW |
| Optimizer AdamW's Learning Rate | 2e-5 |
| Optimizer AdamW's Epsilon | 1e-8 |

Compare to the BERT models with custom finetuning parameters configured, this study was able to use GPT-3 model via its web API with mostly no custom configuration at all. However, the dataset needs to be formatted aligned to the parameters of the OpenAI's GPT-3 API. Columns of the dataset were renamed with the chat from 'message' to 'prompt' and the classification from 'target' to 'completion'. Comparing the models' fine-tuning performance, GPT-3 was able to achieve (0%) training loss and (100%) accuracy on its last iteration step while BERT (Base-uncased) achieved (2%) training loss and (89%) accuracy on its last epoch while the BERT (Large-uncased) attain (14%) training loss and (81%) accuracy on its last epoch.

**Test Predictions**

Twenty-five percent (25%) or 638 of the total 2,552 chat data were used to test the two models. Overall, GPT-3 with eighty-three percent (83%) accuracy was able to achieve the highest accuracy. The BERT (Large-uncased) followed with (82%) accuracy. The BERT (Base-uncased) displays last with eighty percent (80%) accuracy on the toxic chat detection test. BERT (Large-uncased) was able to get the highest ratings in few of the metrics. However, GPT-3 was able to achieve the highest overall performance with the 83% accuracy.

**Figure 3: Displays Precision (a), Recall (b), F1-score (c) and Accuracy (d) of the models on toxic chat detection test**

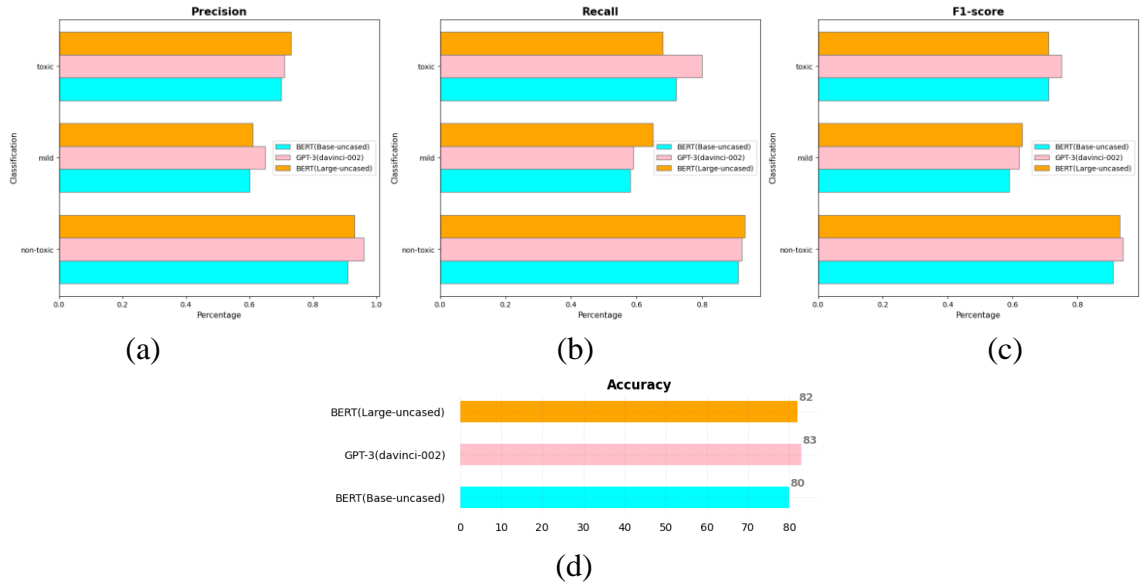

(a)         (b)         (c)

(d)

**Figure 4: Normalized confusion matrix comparison of BERT (Base-uncased) (a), BERT (Large-uncased) (b), and GPT-3 (c) test classification**

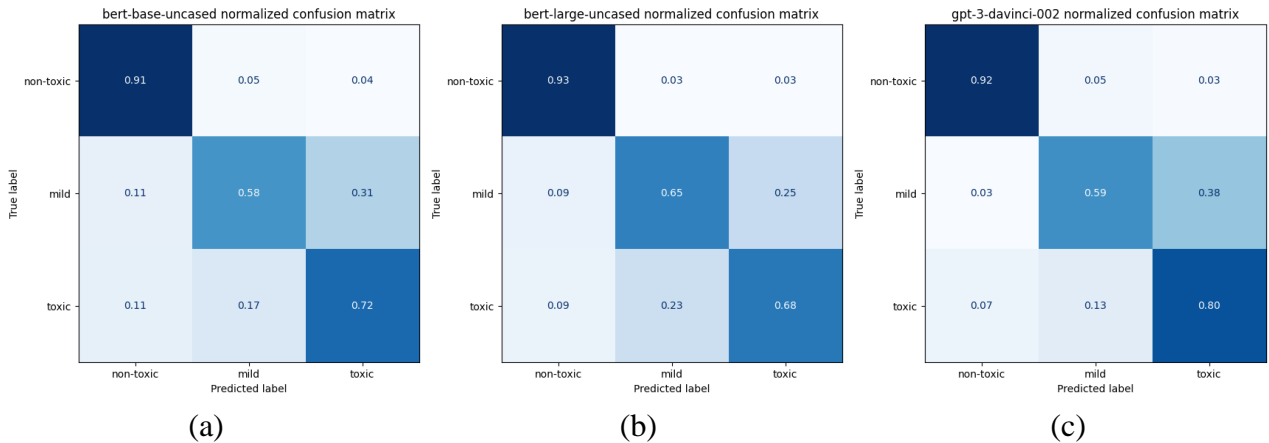

(a)         (b)         (c)

**Table 3: Tabular visualization of classification test performance of the BERT models and GPT-3**

| Classification | Precision | | | Recall | | | F1-score | | |
|---|---|---|---|---|---|---|---|---|---|
| | BERT (base) | BERT (large) | GPT-3 | BERT (base) | BERT (large) | GPT-3 | BERT (base) | BERT (large) | GPT-3 |
| Non-toxic | 0.91 | 0.93 | 0.96 | 0.91 | 0.93 | 0.92 | 0.91 | 0.93 | 0.94 |
| Mild | 0.60 | 0.61 | 0.65 | 0.58 | 0.65 | 0.59 | 0.59 | 0.63 | 0.62 |
| Toxic | 0.70 | 0.73 | 0.71 | 0.72 | 0.68 | 0.80 | 0.71 | 0.71 | 0.75 |

**Table 4: Accuracy performance of the BERT models and GPT-3 model on the test prediction/classification**

| Model | Accuracy |
|---|---|
| BERT (base-uncased) | 0.80 |
| BERT (large-uncased) | 0.82 |
| GPT-3 | 0.83 |

## 5. CONCLUSION

The study successfully collected DOTA 2 in-game chats using OpenDota's public API via web scraping with a Python script. The data was analyzed and in-game chats were classified into three classes naming: non-toxic, mild (toxicity), and toxic. Two BERT models and a GPT-3 model were investigated, fine-tuned, and evaluated to the chat toxicity detection task. Conclusively, the investigation determined that GPT-3 exhibited superior performance compare to the two BERT models in the task of detecting or classifying in-game toxic chats or trash talks. With the state-of-the-art results delivered by the language models, this study finds the promising potential of pre-trained language models. Concluding their effectiveness in mitigating and addressing issues pertaining to toxicity within different online communities and gaming platforms. Building upon these outcomes, possible future works involve extending data collection efforts beyond the DOTA 2 platform to cover a wider array of gaming platforms and online communities. Moreover, other alternative language models and other variations of GPT and BERT can also be a propitious candidate on the task of detecting trash talks and toxic chats.

**References**
1. Lenhart, A., Ybarra, M., Zickuhr, K., & Price-Feeney, M. (2016). Online Harassment, Digital Abuse, and Cyberstalking in America. Data & Society Research Institute. Center for Innovative Public Health Research. https://www.datasociety.net/pubs/oh/Online_Harassment_2016.pdf
2. Unity Multiplayer. (2021). Toxicity in Multiplayer Games Report 2021. The Harris on Demand, The Harris Poll. https://images.response.unity3d.com/Web/Unity/%7Bf8de8201-1c40-4a2a-952a-2468de7b3ce6%7D_Unity_HarrisPoll_ToxicityinGaming.pdf
3. Saarinen, T. 2017. Toxic Behavior in Online Games. University of Oulu, Faculty of Information Technology and Electrical Engineering, Department of Information Processing Science, Information Processing Science.
4. Deloy, Exelsis Deo. (2022). The Culture of Trash Talks Among Dota Players: An Ethnography. International Journal of Research Publications. 109.
5. Lapolla, Matthew, Tackling Toxicity: Identifying and Addressing Toxic Behavior in Online Video Games (2020). Seton Hall University Dissertations and Theses (ETDs). 2798.
6. Meyer, Rachel (2020) Exploring Toxic Behaviour in Online Multiplayer Video Games. MSc by research thesis, University of York.
7. Devlin, J., Chang, M. W., Lee, K., & Toutanova, K. (2018). Bert: Pre-training of deep bidirectional transformers for language understanding. arXiv preprint arXiv:1810.04805.
8. Radford, A., Narasimhan, K., Salimans, T., & Sutskever, I. (2018). Improving language understanding by generative pre-training.
9. Vaswani, A., Shazeer, N., Parmar, N., Uszkoreit, J., Jones, L., Gomez, A. N., ... & Polosukhin, I. (2017). Attention is all you need. Advances in neural information processing systems, 30.

10. Cruz, J. C. B., Tan, J. A., & Cheng, C. (2019). Localization of fake news detection via multitask transfer learning. arXiv preprint arXiv:1910.09295.
11. Rehana, H., Çam, N. B., Basmaci, M., He, Y., Özgür, A., & Hur, J. (2023). Evaluation of GPT and BERT-based models on identifying protein-protein interactions in biomedical text. arXiv preprint arXiv:2303.17728.
12. Radford, A., Wu, J., Child, R., Luan, D., Amodei, D., & Sutskever, I. (2019). Language Models are Unsupervised Multitask Learners.
13. Brown, T., Mann, B., Ryder, N., Subbiah, M., Kaplan, J. D., Dhariwal, P., ... & Amodei, D. (2020). Language models are few-shot learners. Advances in neural information processing systems, 33, 1877-1901.
14. Bach, M., Werner, A., and Palt, M., The Proposal of Undersampling Method for Learning from Imbalanced Datasets, Procedia Computer Science, Volume 159, 2019, Pages 125-134, ISSN 1877-0509.
15. Wolf, T., Debut, L., Sanh, V., Chaumond, J., Delangue, C., Moi, A., ... & Rush, A. M. (2019). Huggingface's transformers: State-of-the-art natural language processing. arXiv preprint arXiv:1910.03771.
16. Blackburn, J., & Kwak, H. (2014, April). STFU NOOB! predicting crowdsourced decisions on toxic behavior in online games. In Proceedings of the 23rd international conference on World wide web (pp. 877-888).
17. Yadav, S., & Kaushik, A. (2023). Comparative Study of Pre-trained Language Models for Text Classification in Smart Agriculture Domain. In Advances in Data-driven Computing and Intelligent Systems: Selected Papers from ADCIS 2022, Volume 2 (pp. 267-279). Singapore: Springer Nature Singapore.
18. Wolf, T., Sanh, V., Chaumond, J., & Delangue, C. (2019). Transfertransfo: A transfer learning approach for neural network based conversational agents. arXiv preprint arXiv:1901.08149.